\documentclass[a4paper]{article}\usepackage[]{graphicx}\usepackage[]{color}
\makeatletter
\def\maxwidth{ %
  \ifdim\Gin@nat@width>\linewidth
    \linewidth
  \else
    \Gin@nat@width
  \fi
}
\makeatother

\definecolor{fgcolor}{rgb}{0.345, 0.345, 0.345}

\usepackage{framed}
\makeatletter
 {\par\unskip\endMakeFramed%
 \at@end@of@kframe}
\makeatother

\definecolor{shadecolor}{rgb}{.97, .97, .97}
\definecolor{messagecolor}{rgb}{0, 0, 0}
\definecolor{warningcolor}{rgb}{1, 0, 1}
\definecolor{errorcolor}{rgb}{1, 0, 0}
\newenvironment{knitrout}{}{} % an empty environment to be redefined in TeX

\usepackage{alltt}

\usepackage[utf8]{inputenc}
\usepackage{hyperref}
\usepackage{authblk}
\IfFileExists{upquote.sty}{\usepackage{upquote}}{}
\begin{document}

\title{There is no fast lunch: an examination of the running speed of
  evolutionary algorithms in several languages}

\author[1]{J.J. Merelo\thanks{{\tt jmerelo@ugr.es}. Reachable also at the
    \href{https://github.com/geneura-papers/2015-ea-languages/issues}{issues 
section}
    of the repository for this paper.}}
\author[1]{P. Garc\'ia-S\'anchez \thanks{fergunet@gmail.com}}
\author[2]{M. Garc\'ia-Valdez\thanks{mariosky@gmail.com}}
\author[1]{I. Blancas\thanks{iblancasa@gmail.com}}
      
\affil[1]{University of Granada, Spain}
\affil[2]{Instituto Tecnológico de Tijuana, Mexico}

\date{}

\maketitle

\begin{abstract}
  It is quite usual when an evolutionary algorithm tool or library
  uses a language other than C, C++, Java or Matlab that a reviewer or
  the audience questions its usefulness based on the speed of those
  other languages, purportedly slower than the aforementioned
  ones. Despite speed being not everything needed to design a useful
  % Pablo: confirmado en Word Reference, es A useful (se pronuncia
  % yiuseful) 
  
  evolutionary algorithm application, in this paper we will measure
  the speed for several very basic evolutionary algorithm operations
  in several languages which use different virtual machines and %Pablo: poner "scripting languages" para dejarlo claro?
  approaches, and prove that, in fact, there is no big difference in
  speed between interpreted and compiled languages, and that in some
  cases, interpreted languages such as JavaScript or Python can be faster than compiled languages
  such as Scala, making them worthy of use for evolutionary algorithm
  experimentation.  
\end{abstract}

\section{Introduction}

It is a well extended myth in scientific programming to claim that
compiled languages such as C++ or Java are always, in every
circumstance, faster than interpreted languages such as Perl,
JavaScript or Python.

However, while it is quite clear that efficiency matters, as said in
 \cite{anderson2010efficiency}, in general and restricting the concept
 of {\em speed} to {\em speed of 
  the compiled/interpreted application} it might be the case that some
languages are faster to others, as evidenced by benhmarks such as
\cite{prechelt2000empirical,fulghamcomputer}. Taken in general or even
restricting it to some particular set of problems such as floating
point computation, some compiled languages tend to be faster than
interpreted languages.

But, in the same spirit of the {\em There is no free lunch} theorem
\cite{Wolpert-1997-NFL} we can affirm there is a {\em no fast lunch}
theorem for the implementation of evolutionary optimization, in the
sense that, while there are particular languages that might be the
fastests for particular problem sizes and specially fitness functions,
in general the fastest language will have those two dependencies, and,
specially, for non-trivial problem sizes and limiting ourselves to the
realm of evolutionary algorithm operators, scripting languages such as
JavaScript might be as fast or even faster than compiled languages
such as Java.

Coming up next, we will write a brief state of the art of the analysis
of implementations of evolutionary algorithms. Next we will present
the test we have used for this paper and its rationale, and finally we
will present the results of examining four different languages running
the most widely used evolutionary algorithm operator:
mutation. Finally, we will draw the conclusions and present future
lines of work. 

\section{State of the art}

In fact, the examination of the running time of an evolutionary
algorithm has received some attention from early on. Implementation
matters \cite{DBLP:conf/iwann/MereloRACML11,nesmachnow2011time}, which implies that
paying attention to the particular way an algorithm is implemented
might result in speed improvements that outclass that achieved by %Mario: those achieved (improvements)
using the {\em a priori} fastest language available. In fact, careful
coding led us to prove \cite{ae09} that Perl,
an interpreted and not optimized for speed language, could obtain times
that were on the same order the magnitude as Java. However, that 
paper also proved that, for the particular type of problems used in
scientific computing in general, the running speed is not as important
as coding speed or even learning speed, since most scientific programs
are, in fact, run a few times while a lot of time is spent on coding
them. That is why expressive languages such as Perl, JavaScript or
Python are, in many cases, superior to these fast-to-run
languages. %Pablo: bueno, algún reviewer podría decirte que hay
           %frameworks ya hechos 
% Es un tech report.

However the benchmarks done in those papers were restricted to
particular problem sizes. Since program speed is the result of many
factors, including memory management and implementation of loop control %Pablo: Esta frase deberías haberla puesto en la intro también
%Mario: No sé si loop data structure es muy utilizado, nunca lo había escuchado. 
%Cambiado a control structure.
structures, in this paper we will examine how fast several languages
are for different problem sizes. This will be done next. 

\section{Experimental setup}

First, a particular problem was chosen for testing different
languages and also data representations: performing bit-flip mutation
on a binary string. In fact, this is not usually the part of the
program an evolutionary algorithm spends the most time in
\cite{nesmachnow2011time}. In general, that is the fitness function,
and then reproduction-related functions: chromosome ranking, for
instance. However, mutation is the operation that is performed the
%Mario: Yo creo que este operador de mutación no es el que más se realiza 
%ya que su aplicación tiene una probabilidad.
%Bueno, quizás menos que crossover. Para el siguiente paper...
most times on every evolutionary algorithm and is quintessential to
the algorithm itself, so it allows the comparison of the different
languages in the proper context. 

Essentially, mutation is performed by \begin{enumerate}

\item Generating a random integer from 0 to the length of the chromosome. %Pablo: Generating
\item Choosing the bit in that position and flipping it
\item Building a chromosome with the value of that bit changed.

\end{enumerate}

Chromosomes can be represented in at least two different ways: an
array or vector of boolean values, or any other scalar value that can
be assimilated to it, or as a bitstring using generally ``1'' for true
values or ``0'' for false values. Different data structures will have
an impact on the result, since the operations that are applied to them
are, in many cases, completely different and thus the underlying
implementation is more or less efficient.

Then, four languages have been chosen for performing the
benchmark. The primary reason for chosing these languages was the
availability of open source implementations for the author, but also
they represent different philosophies in language design.

\begin{table}[htb]
    \centering
    \begin{tabular}{l|c|l}
      \hline
      Language & Version & URL \\
      \hline
      Scala & 2.11.7 & \url{http://git.io/bfscala} \\
      Lua & 5.2.3 & \url{http://git.io/bflua} \\
      Perl & v5.20.0 & \url{http://git.io/bfperl} \\
      JavaScript & node.js 5.0.0 & \url{http://git.io/bfnode} \\ http://git.io/bfpython
      Python & 2.7.3 & \url{http://git.io/bfpython} \\ 
      \hline 
      \end{tabular}
      \caption{Languages used and file written to carry out the
        benchmark. No special flags were used for the interpreter or
        compiler. \label{tab:files}}
    \end{table}
Compiled languages are represented by Scala, a strongly-typed %Pablo: no se quejarán de que no hayamos usado Java directamente? También: presented o represented?
% Puede... para el siguiente paper
functional language that compiles to a Java Virtual Machine 
%Mario:that compiles to a Java Virtual Machine bytecode ?
bytecode. Scala is in many cases faster than Java
%fixed
\cite{fulghamcomputer} due to its more efficient
implementation of type handling. Two different representations were
used in Scala: {\tt String} and {\tt Vector[Boolean]}. They both have
the same underlying type, {\tt IndexedSeq} and in fact the overloading
of operators allows us to use the same syntax independently of the
type. The benchmark, {\tt bitflip.scala}, is available under a GPL
license, at the URL shown in Table \ref{tab:files}. 

Interpreted languages are represented by Lua, Perl and Javascript. Lua
is a popular embedded language that is designed for easy
implementation; Perl has been used extensively for evolutionary
algorithms \cite{ae09,merelo14:noisy,DBLP:conf/cec/GuervosMCCV13} with
satisfactory results, and node.js, an implementation of JavaScript,
which uses a the V8 JIT compiler to create bytecode when in reads the
script, and 
% Mario: and node.js, a JavaScript framework
% la implementación que usa es Google V8. 
% Según: https://developers.google.com/v8/design
% V8 compiles JavaScript source code directly into machine code when it is first executed.   
has been used lately by our research group as part of our NodEO
library \cite{DBLP:conf/gecco/GuervosVGES14} and volunteer computing
framework NodIO \cite{DBLP:journals/corr/GuervosG15}. In fact, this 
paper is in part a rebuttal to concerns made by reviewers of the lack of %Mario: No eran claims eran concerns :)
speed and thereof adequacy of JavaScript for evolutionary algorithm
% Pablo: está bien decir esto? :P JJ: Lo estamos haciendo...
experimentation. Versions and files are shown in the Table
\ref{tab:files}. Only Perl used two data structures as in Scala: a
string, which is a scalar structure in Perl, and an array of
booleans. 

In all cases except in Scala, implementation took less than one hour and was
inspired by the initial implementation made in Perl. Adequate data and control
structures were used for running the application, which applies
mutation to a single generated chromosome a hundred thousand
times. The length of the mutated string starts at 16 and is doubled
until $2^{15}$ is reached, that is, 32768. This upper length was chosen to
have an ample range, but also so small as to be able to run the
benchmarks within one hour. Results are shown next.

\section{Results and analysis}
\label{sec:res}

\begin{figure}[h!tb]
  \centering
\begin{knitrout}
\definecolor{shadecolor}{rgb}{0.969, 0.969, 0.969}\color{fgcolor}
\includegraphics[width=\maxwidth]{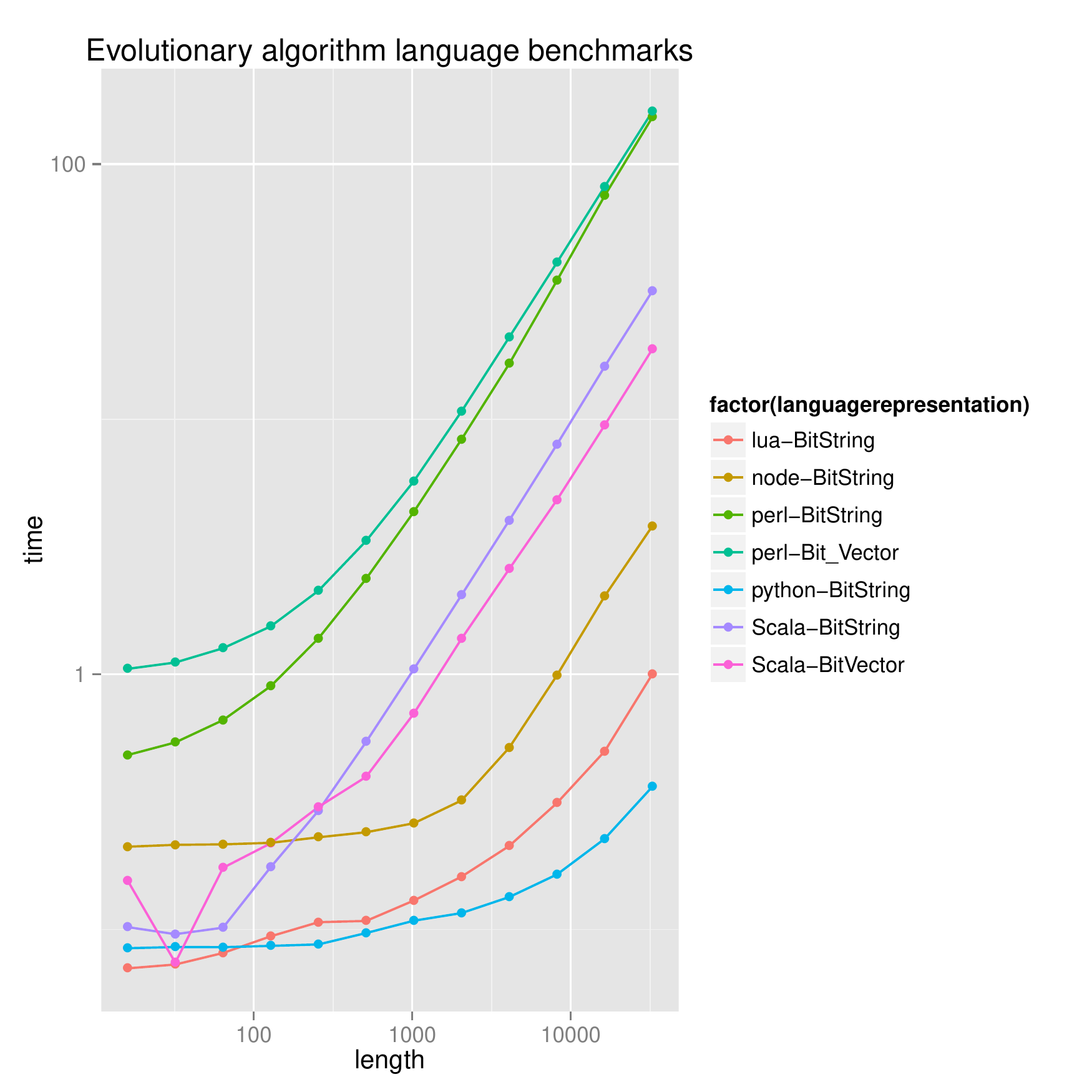} 

\end{knitrout}
\caption{Plot of time needed to perform 100K mutations in strings with
lengths increasing by a factor of two from 16 to $2^{15}$. Please note
that $x$ and $y$ both have a logarithmic scale.}
\label{fig:time}
\end{figure}

All measurements and processing scripts are included in this paper
repository, although in fact the programs were written to directly
produce a CSV (comma separated value) representation of measurements,
which was then plotted using R and {\tt ggplot} as shown in Figure
\ref{fig:time}. The first unexpected behavior shown here is the
remarkable speed of the Lua language, which is, in fact, faster
than any other small sizes, although slower than Python at bigger
sizes. After these two, next position goes to node.js, which uses a very
efficient implementation of a Just-in-Time interpreter for the %Mario: Si se usó V8 es un compilador 
% Es un intérprete, porque se lee cada vez del fuente. - jj
JavaScript language. Then Scala, whose bitvector representation is
better than the bitstring and finally Perl, with a bitstring representation being
slightly better than bit vectors, although the difference dilutes with
time. In fact, Scala is a bit better than node for the
smaller sizes, less than 128 bits and any representation, but that advantage disappears for greater sizes. 

The behavior of the languages is not linear either. Node.js and Python
both have an
interesting feature: its speed is roughly independent of the string
size up to size 1024. The same happens also for several size
segments for Lua, showing some plateaus that eventually break for the
bigger sizes. It probably means that the creation and accessing of strings is
done in constant time, or is roughly constant, giving it a performance
advantage over other languages. Even so, it never manages to beat the
fastest language in this context, which is either Lua or Python,
depending on the size. 

The trend from size 1024 on is for the differences to keep in more or less the same   
style for bigger sizes, so we do not think it would be interesting to
extend it to $2^16$ and upwards. In any case, these measures allow us %Pablo: confirmado en WR, In any.
to measure the performance of the most widely used genetic operator in
four different and popular languages, since all four of them (except
for Lua) show up in most rankings of the most popular languages, such
as the Tiobe ranking \cite{tiobe15}.  %Pablo: algún enlace?
%Mario: Un ranking popular es: http://www.tiobe.com/index.php/content/paperinfo/tpci/index.html

\section{Conclusions}

In this paper we set out to measure the speed of different languages
when running the classical evolutionary algorithm operation: mutating
a chromosome represented as a binary string or vector. The results
can be a factor on the choice of a language for implementing solution
to problems using evolutionary algorithms, at least if raw running
speed is the main consideration. And, if that is our main concern, the
fastest language for mutating strings has been found to be Lua and Python,
followed by node.js and Perl. Most interpreted languages are faster for the
wider range of chromosome sizes than Scala, which is a compiled
language that uses the Java Virtual machine. 

Despite its speed Lua is not exactly a popular language, although it definitely
has found its niche in embedded systems and applications such as in-game scripting,
game development or servers so 
our choice for EA programming languages would be Python or JavaScript,
which are fast enough, popular, allow for fast development and have a
great and thriving community. JavaScript does have an advantage over
Python: Besides being interpreted languages and using
dynamic typing, it can express complex operations in a terse syntax
and bestows implementations both in browsers and on the
server. We can conclude from these facts and the measurements made in
this paper that JavaScript is perfectly adequate for any scientific
computing task, including evolutionary algorithms.

That does not mean that Perl or Scala  are not adequate for
scientific computing. However, they
might not be if experiments take a long time and time is of the
essence; in that case implementing the critical parts of the program
using C, Go, Lua or Python might be the right way to go. But in general, it
cannot be said that interpreted languages are not an adequate platform
for implementing evolutionary algorithms, as proved in this paper. %Pablo: esto sí mola ponerlo

Future lines of work might include a more extensive measurement of
other operators such as crossover, tournament selection and other
selection algorithms. However, they are essentially CPU integer
operations and their behavior might be, in principle, very similar to %Pablo: arriba behaviour estaba en americano
the one shown here. This remains to be proved, however, but it is left
as future line of work.

\section{Acknowledgements}

This paper is part of the open science effort at the university of
Granada. It has been written using {\tt knitr}, and its source as well as
the data used to create it can be downloaded from
\href{https://github.com/geneura-papers/2015-ea-languages}{the GitHub
  repository}. It has been supported in part by  
\href{http://geneura.wordpress.com}{GeNeura Team}. 

This work has been supported in part SPIP2014-01437 (Direcci\'on General
de Tr\'afico), PRY142/14 (Fundaci\'on P\'ublica Andaluza Centro de
Estudios Andaluces en la IX Convocatoria de Proyectos de
Investigaci\'on), TIN2014-56494-C4-3-P (Spanish Ministry of Economy
and Competitivity), and PYR-2014-17 GENIL project (CEI-BIOTIC
Granada).

\bibliographystyle{elsarticle-num}
\bibliography{geneura,languages,GA-general}

\end{document}